\pdfoutput=1

\documentclass[11pt,a4paper]{article}

\usepackage{acronym}
\usepackage{amsmath}
\usepackage[hyperref]{acl2021}
\usepackage{times}
\usepackage{latexsym}
\usepackage{microtype}
\usepackage{graphicx}
\usepackage[ruled,vlined, linesnumbered]{algorithm2e}
\usepackage{booktabs}
\usepackage[inline]{enumitem}
\usepackage{amssymb}
\usepackage[skip=0pt]{caption}
\usepackage{booktabs}
\usepackage{makecell}
\usepackage{multirow}

\SetKwInput{KwInput}{Input} 
\SetKwInput{KwOutput}{Output}

\acrodef{CIS}{Conversational Information Seeking}
\acrodef{CQS}{Conversational Question Simplification}
\acrodef{CQR}{Conversational Question Rewriting}
\acrodef{CQA}{Conversational Question Answering}
\acrodef{LD}{Levenshtein distance}
\acrodef{LM}{Language Model}
\acrodef{DP}{Dynamic Programming}
\acrodef{RL}{Reinforcement Learning}
\acrodef{MLE}{maximum likelihood estimation}
\acrodef{MLD}{minimum Levenshtein distance}
\acrodef{EMLD}{edit-based minimum Levenshtein distance}
\acrodef{Bert}{Bidirectional Encoder Representations from Transformers}
\acrodef{RISE}{Reinforcement Iterative Sequence Editing}
\acrodef{K}{\textbf{K}eep}
\acrodef{I}{\textbf{I}nsert}
\acrodef{D}{\textbf{D}elete}
\acrodef{S}{\textbf{S}ubstitute}
\acrodef{CANARD}{Context Abstraction: Necessary Additional Rewritten Discourse}
\acrodef{MDP}{Markov Decision Process}
\acrodef{HCMDP}{Hierarchical Combinatorial Markov Decision Process}
\acrodef{CMDP}{Composite Markov Decision Process}
\acrodef{SMDP}{Semi-Markov Decision Process}
\acrodef{TD}{Temporal difference}
\acrodef{IRT}{Iterative Reinforce Training}
\acrodef{DPS}{Dynamic Programming based Sampling}
\acrodef{RNN}{Recurrent neural network}
\acrodef{QA}{Question Answering}
\acrodef{POS}{Part-Of-Speech}
\acrodef{Seq2Seq}{sequence-to-sequence}
\acrodef{NS}{Nucleus Sampling}
\acrodef{NLI}{Natural Language Inference}
\acrodef{BE}{Boltzmann Exploration}
\acrodef{CAsT}{Conversational Assistance Track}
\acrodef{GAN}{Generative Adversarial Network}

\DeclareMathOperator{\softmax}{softmax}

\allowdisplaybreaks

\aclfinalcopy

\title{Learning to Ask Conversational Questions \\ by Optimizing Levenshtein Distance}

\author{Zhongkun Liu\textsuperscript{\rm 1}, \  Pengjie Ren\textsuperscript{\rm 1}\thanks{$^*$ Corresponding authors.} , \ Zhumin Chen\textsuperscript{\rm 1}$^*$, \ Zhaochun Ren\textsuperscript{\rm 1}, \\ \bf Maarten de Rijke\textsuperscript{\rm 2}, \ Ming Zhou\textsuperscript{\rm 3}  
\\
  \textsuperscript{\rm 1}School of Computer Science and Technology, Shandong University, China \\
  \textsuperscript{\rm 2}University of Amsterdam \& Ahold Delhaize \\
  \textsuperscript{\rm 3}Sinovation Ventures, China \\
  \texttt{\{liuzhongkun,renpengjie,chenzhumin, zhaochun.ren\}@sdu.edu.cn} \\
  \texttt{m.derijke@uva.nl}; \  \texttt{mingzhou926@hotmail.com}
  
  }

\date{}

\begin{document}

\maketitle


\begin{abstract}
\acf{CQS} aims to simplify self-contained questions into conversational ones by incorporating some conversational characteristics, e.g., anaphora and ellipsis.
Existing \acl{MLE} based methods often get trapped in easily learned tokens as all tokens are treated equally during training.
In this work, we introduce a \acf{RISE} framework that optimizes the \acl{MLD} through explicit editing actions.
\ac{RISE} is able to pay attention to tokens that are related to conversational characteristics.
To train \acs{RISE}, we devise an \acf{IRT} algorithm with a \acf{DPS} process to improve exploration.
Experimental results on two benchmark datasets show that \acs{RISE} significantly outperforms state-of-the-art methods and generalizes well on unseen data.
\end{abstract}


\section{Introduction}
        
        %

Conversational information seeking (\acs{CIS}\acused{CIS})~\citep{zamani2020macaw,wise} has received extensive attention. 
It introduces a new way to connect people to information through conversations~\citep{open_retrieval,gao2021advances,RenCM0R20}.
One of the key features of \ac{CIS} is \emph{mixed initiative} behavior, where a system can improve user satisfaction by proactively asking clarification questions~\citep{zhang2018towards,sigir/AliannejadiZCC19,conf/emnlp/XuWTDYZZS19}, besides passively providing answers~\citep{croft2010search,chiir_RadlinskiC17,lei2020estimation}.

\begin{figure}[t]
    \centering
    \includegraphics[width=0.48\textwidth]{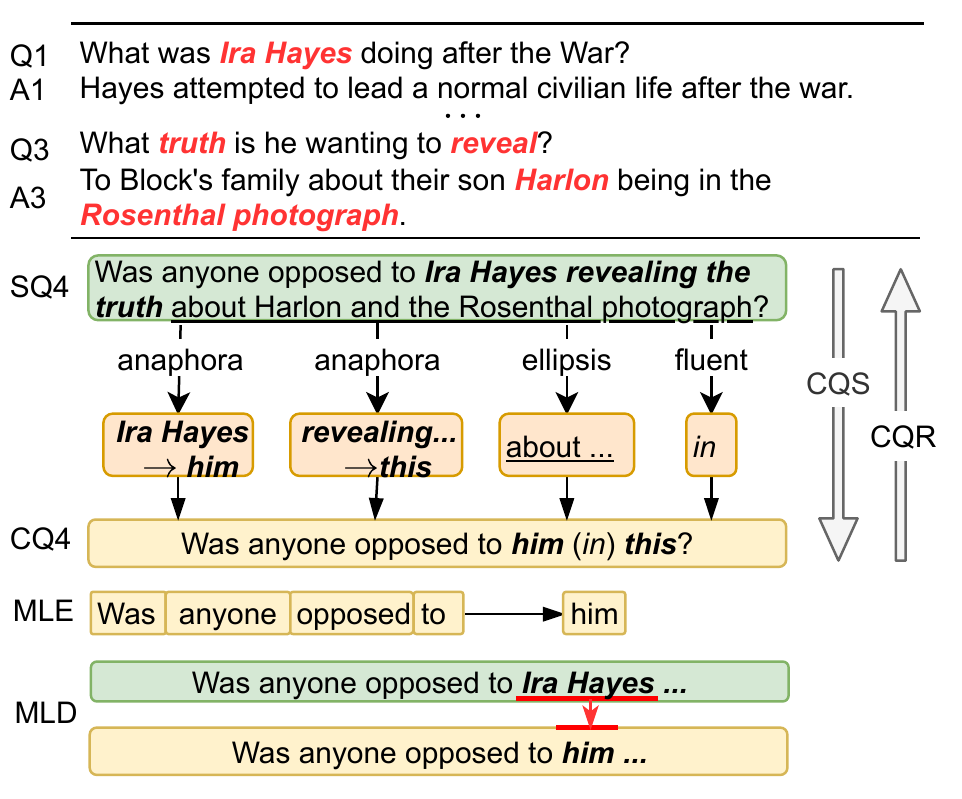}
    \caption{An example for \acl{CQS} and its reverse, \acl{CQR}. 
    Q1--A3 is the context, SQ4 is the self-contained question, and CQ4 is the conversational question. 
    }
    \label{fig:intro_example}
\end{figure}

Previous studies on asking clarification questions can be grouped into two categories: conversational question generation~\citep{conf/emnlp/DuanTCZ17} and conversational question ranking~\citep{sigir/AliannejadiZCC19}.
The former directly generates conversational questions based on the dialogue context.
However, the generated questions may be irrelevant and meaningless~\citep{www/RossetXSCCTB20}. 
A lack of explicit semantic guidance makes it difficult to produce each question token from scratch while preserving relevancy and usefulness at the same time~\citep{conf/acl/HuangNWL18,conf/acl/ChaiW20}.
Instead, the latter proposes to retrieve questions from a collection for the given dialogue context, which can usually guarantee that the questions are relevant and useful~\cite{shen2018knowledge,www/RossetXSCCTB20}.
However, question ranking methods do not lead to a natural communication between human and machine~\citep{pulman1995anaphora}, as they neglect important characteristics in conversations, e.g., anaphora and ellipsis.
As shown in Fig.~\ref{fig:intro_example}, the self-contained question (SQ4) lacks these characteristics, which makes it look unnatural.

In this work, we study the task of \acfi{CQS}.
Given a dialogue context and self-contained question as input, \ac{CQS} aims to transform the self-contained question into a conversational one by simulating conversational characteristics, such as anaphora and ellipsis.
For example, in Fig.~\ref{fig:intro_example}, four simplification operations are applied to obtain the conversational question (CQ4), which is context-dependent and superior to its origin one (SQ4) in terms of naturalness and conveying.
The reverse process, i.e., \ac{CQR}~\citep{elgohary2019can,VoskaridesLRKR20} which rewrites CQ4 into SQ4, has been widely explored in the literature~\cite{vakulenko2020question,conf/sigir/YuLYXBG020}.
Although the proposed methods for \ac{CQR} can be easily adopted for \ac{CQS}, they do not always generate satisfactory results as they are all trained to optimize a \acf{MLE} objective, which gives equal attention to generate each question token.
Therefore, they often get stuck in easily learned tokens, i.e., tokens appearing in input, ignoring conversational tokens, e.g., him, which is a small but important portion of output.

To address the above issue, we propose a new scheme for \ac{CQS}, namely \acfi{MLD}.
It minimizes the differences between input and output, forcing the model to pay attention to contributing tokens that are related to conversational tokens, e.g., ``Ira Hay'' and ``him'' in Fig.~\ref{fig:intro_example}.
Therefore, \ac{MLD} is expected to outperform \ac{MLE} for \ac{CQS}.
However, \ac{MLD} cannot be minimized by direct optimization due to the discrete nature, i.e., minimizing the number of discrete edits.
We present an alternative solution, a \textbf{R}einforcement \textbf{I}terative \textbf{S}equence \textbf{E}diting (\acs{RISE}\acused{RISE}) framework for the optimization of \ac{MLD}.

We formulate \ac{RISE} as a \acf{HCMDP} consisting of an editing \ac{MDP} to predict multiple edits for all tokens in the self-contained question, e.g., `\ac{K}' to keep a token, and a phrasing \ac{MDP} to predict a phrase if the edit is `\ac{I}' or `\ac{S}'.
We only have the self-contained and conversational question pairs in the dataset while the demonstrations of the editing iterations are lacked.
Thus, we cannot train each editing iteration of \ac{RISE} with teacher forcing.
To this end, we devise an \acf{IRT} algorithm that allows \ac{RISE} to do some exploration itself.
The exploration can be rewarded according to its \ac{LD} with the demonstrated conversational question.
Traditional exploration methods like $\epsilon$-sampling~\cite{books/lib/SuttonB98} neglect the interdependency between edits for all tokens, resulting in poor exploration.
Thus, we further introduce a \acf{DPS} process that adopts a \ac{DP} algorithm to track and model the interdependency in \ac{IRT}.
Experiments on the \acs{CANARD}~\cite{elgohary2019can} and \acs{CAsT}~\cite{dalton2019cast} datasets show that \acs{RISE} significantly outperforms state-of-the-art methods and generalizes well to unseen data.







\section{\acl{CQS}: From \acl{MLE} to \acl{MLD}}



\subsection{\acs{CQS}}
Given a dialogue context $C$ representing the previous conversation utterances and the self-contained clarification question candidate $x=\{x_1, \ldots, x_{|x|}\}$ to be asked next (e.g., from a conversational question ranking model), the goal of \acf{CQS} is to reformulate question $x$ to a conversational question $y=\{y_1, \ldots, y_{|y|}\}$ by simulating conversational characteristics, e.g., anaphora and ellipsis.
A target conversational question $y^*=\{y^*_1, \ldots, y^*_{|y^*|}\}$ is provided during the training phase.
    
\subsection{\Acl{MLE} for \acs{CQS}}
A commonly adopted paradigm for tasks similar to \ac{CQS}, e.g., \ac{CQR}, is to model the task as a conditional sequence generation process parameterized by $\theta$, which is usually optimized by \ac{MLE}:
\begin{equation}
    \label{MLE}
    \begin{split}
        \mathcal{L}_\theta &= -\log p_\theta(y^*|x, C) \\
                    &= -\sum_{t=1}^{|y^*|} \log p_\theta(y_t^*|y^*_{<t}, x, C),
    \end{split}
\end{equation}
where $y^*$ is the target question and $y^*_{<t}$ denotes the prefix $y^*_1, y^*_2, \ldots, y^*_{t-1}$.
As we can see, \ac{MLE} gives equal weight to each token and falls in easily learned tokens, the overwhelming duplicate tokens between $x$ and $y$, while underestimating subtle differences of tokens related to conversational characteristics.

\subsection{\Acl{MLD} for \ac{CQS}}
Inspired by \citet{journals/corr/ArjovskyCB17}, to minimize the distance between two distributions, we propose to minimize the \ac{LD} between the target question $y^*$ and the model output $y$ so as to leverage the high overlap between $x$ and $y$ and focus on subtle different tokens:
\begin{equation}
    \label{LD}
    \mathcal{L}_\theta = LD(y, y^*).
\end{equation}
Unfortunately, it is impossible to directly optimize Eq.~\ref{LD} because the \ac{LD} between $y$ and $y^*$ is the minimum number of single-token edits (insertions, deletions or substitutions) required to change $y$ into $y^*$, which is discrete and non-differentiable.

\section{\ac{RISE}}
To optimize \ac{MLD} in Eq.~\ref{LD}, we devise the \acf{RISE} framework, which reformulates the optimization of \ac{MLD} as a \acf{HCMDP}.
Next, we first describe our \ac{HCMDP} formulation of \ac{RISE}.
We then detail the modeling of each ingredient in \ac{RISE}.
Finally, we present the training process of \ac{RISE}.

\subsection{\acs{HCMDP} formulation for \acs{RISE}} \label{Formulation}
\acs{RISE} produces its output $y$ by iteratively editing $x$  with four types of edit, i.e., `\ac{K}' to keep a token, `\ac{D}' to delete a token, `\ac{I}' to insert a phrase~(a sequence of tokens) after a token, and `\ac{S}' to substitute a phrase by a new one.
If a token is predicted as `\ac{I}' or `\ac{S}', we need to further predict a corresponding phrase.
Note that we only predict one phrase for successive `\ac{S}' edits.
We formulate \ac{RISE} as a \acf{HCMDP} consisting of 
\begin{enumerate*}[label=(\arabic*)]
\item an \emph{editing} \ac{MDP} to predict multiple edits for all tokens, and 
\item a \emph{phrasing} \ac{MDP} to predict a phrase if the edit is `\ac{I}' or `\ac{S}'.
\end{enumerate*}

The editing \acs{MDP} can be formulated as a tuple $\langle\mathcal{S}^e, \mathcal{A}^e, \mathcal{T}^e, \mathcal{R}, \pi^e \rangle$. 
Here, $s^e_t \in \mathcal{S}^e$ denotes the question at $t$-th iteration $y^t$ together with the context $C$, i.e., $s^e_t=(y^t, C)$.
Note that $s^e_0=(x, C)$.
$a^e_t=[a^e_{t, 1}, a^e_{t, 2}, \ldots, a^e_{t, |y^t|}] \in \mathcal{A}^e$ is a combinatorial action consisting of several interdependent edits. 
The number of edits corresponds to the length of $y^t$.
For example, in Fig.~\ref{fig:model}, $a^e_t =$ [`\ac{K}', `\ac{K}', `\ac{K}', `\ac{K}', `\ac{S}', `\ac{S}', `\ac{K}', `\ac{K}'].
In our case, the transition function $\mathcal{T}^e$ is deterministic, which means that the next state $s^e_{t+1}$ is obtained by applying the predicted actions from both the editing \ac{MDP} and phrasing \ac{MDP} to the current state $s^e_t$.
$r_t \in \mathcal{R}$ is the reward function, which estimates the joint effect of taking the predicted actions from both the editing and phrasing \acp{MDP}.
$\pi^e$ is the editing policy network.

The phrasing \acs{MDP} can be formulated as a tuple $\langle\mathcal{S}^p, \mathcal{A}^p, \mathcal{T}^p, \mathcal{R}, \pi^p \rangle$. 
Here, $s^p_{t} \in \mathcal{S}^p$ consists of the current question $y^t$, the predicted action from the editing \ac{MDP} $a^e_t$, and the context $C$, i.e., $s^p_{t}=(y^t, a^e_t, C)$.
$a^p_t = [a^p_{t, 1}, a^p_{t, 2}, \ldots] \in \mathcal{A}^p$ is also a combinatorial action, where $a^p_{t, i}$ denotes a phrase from a predefined vocabulary and $i$ corresponds to the index of the `\ac{I}' or `\ac{S}' edits, e.g., in Fig.~\ref{fig:model}, `$a^p_{t, 1} = \text{him}$' is the predicted phrase for the first `\ac{S}' edit.
The length of the action sequence corresponds to the number of `\ac{I}' or `\ac{S}' edits.
The transition function $\mathcal{T}^p$ returns the next state $s^p_{t+1}$ by applying the predicted actions from the phrasing \ac{MDP} to the current state $s^p_t$.
$r_t \in \mathcal{R}$ is the shared reward function.
$\pi^p$ is the phrasing policy network.

\ac{RISE} tries to maximize the expected reward:
\begin{equation}
\label{OptFunc}
 J(\theta) = E_{a^e_t \sim \pi^e, a^p_t \sim \pi^p} [r_t],
\end{equation}
where $\theta$ is the model parameter which is optimized with the policy gradient:
\begin{equation}
    \label{PG}
    \begin{split}
    \mbox{}\hspace*{-2mm}
        \nabla J(\theta) = E_{a^e_t \sim \pi^e, a^p_t \sim \pi^p} [r_t (&\nabla \log \pi^e(a^e_t|s^e_t) +
            \hspace*{-2mm} \mbox{} \\ 
            &\nabla \log \pi^p(a^p_t|s^p_t))], \hspace*{-2mm} \mbox{}
    \end{split}
\end{equation}
Next, we will show how to model $\pi^e(a^e_t|s^e_t)$, $\pi^p(a^p_t|s^p_t)$, and $r_t$.

\subsection{Policy networks}
\begin{figure*}[ht]
    \centering
    \includegraphics[width=\textwidth]{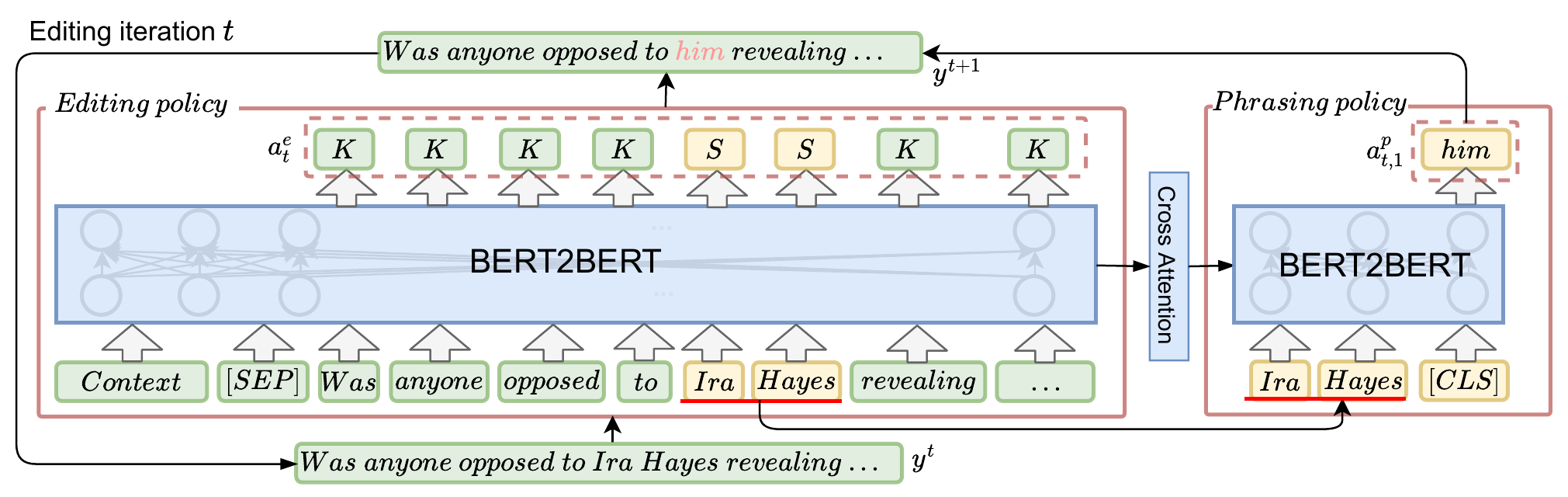}
    \caption{Architecture of our policy network.
            A combinatorial of all tokens edits is predicted by editing policy, and for each `\ac{I}' or `\ac{S}' edit, a phrase will be predicted by phrasing policy.}
    \label{fig:model}
\end{figure*}
We implement the editing and phrasing policy networks ($\pi^e$ and $\pi^p$) based on BERT2BERT~\citep{journals/tacl/RotheNS20} as shown in Fig.~\ref{fig:model}. 
The editing policy network is implemented by the encoder to predict combinatorial edits, and the phrasing policy network is implemented by the decoder to predict phrases.

\subsubsection{Editing policy network}
We unfold all tokens of the utterances in the context into a sequence $C=(w_1, \ldots, w_c)$, where $w_i$ denotes a token and we add ``[SEP]'' to separate different utterances.
Then the context and input question in $t$-th iteration are concatenated with ``[SEP]'' as the separator.
Finally, we feed them into the encoder of BERT2BERT to obtain hidden representations for tokens in question $H^t=(h^t_1, \ldots, h^t_{|y^t|})$ and apply a linear layer with parameter $W^e$ to predict $a^e_t$:
\begin{equation}
    \label{edit_prediction}
    \pi^e(a^e_t|s^e_t=(y^t, C)) = \softmax(W^e H^t).
\end{equation}

\subsubsection{Phrasing policy network}
We first extract the spans corresponding to the `\ac{I}' or `\ac{S}' edits from the question.
If the edit is `\ac{I}', the question span $span^{t}_i$ consists of tokens before and after this insertion, i.e., $span^{t}_i =[y^t_j, y^t_{j + 1}]$;
if the edit is `\ac{S}', the question span $span^{t}_i$ consists of successive tokens corresponding to the `\ac{S}' edit, i.e., $span^{t}_i=[y^t_j, \ldots, y^t_k]$, where $a^e_{t, j:k}=$`\ac{S}' and $a^e_{t, k + 1} \neq$ `\ac{S}'. 
We only predict once for successive `\ac{S}' edits, e.g., in Fig.~\ref{fig:model}, the phrase `him' is predicted to substitute question span [``Ira'', ``Hayes''].

For the $i$-th `\ac{I}' or `\ac{S}' edit with a question span $span^{t}_i$, we concatenate the span and ``[CLS]'' token as input tokens, and feed them into the decoder of BERT2BERT to obtain a hidden representation of ``[CLS]'' token $s^t_i$.
We obtain $S^t$ by concatenating each $s^t_i$ and predict the phrases for all `S' and `I' edits by a linear layer with parameter $W^p$:
\begin{equation}
    \label{phrase_prediction}
    \pi^p(a^p_t|s^p_t) = \softmax(W^p S^t).
\end{equation}

\subsection{Reward $R$}
We devise the reward $r_t$ to estimate the effect of taking the joint action $(a^e_t, a^p_t)$ by encouraging actions that can result in low \ac{LD} values between $y^{t + 1}$ and $y^*$, i.e., minimizing Eq.~\ref{LD}.
Besides, we discourage those actions to achieve same $y^{t + 1}$ with extra non `K' edits:
\begin{equation}
    \label{reward}
    \begin{split}
        r_t ={}& \frac{1}{1 + LD(y^{t + 1}, y^*)} \times \\
             & \left(l - \sum_t (a^e_t \neq \text{`K'}) + 1\right), \\
         l = {} & LD(y^{t}, y^*) - LD(y^{t + 1}, y^*),
    \end{split}
\end{equation}
where $\frac{1}{1 + LD(y^{t + 1}, y^*)}$ will reward actions that result in low \ac{LD} values between $y^{t + 1}$ and $y^*$ and $(l - \sum_t (a^e_t \neq \text{`K'}))$ will punish those actions with unnecessary non `K' edits.

\subsection{Training}
To train \ac{RISE}, we need training samples in the form of a tuple $(s_t^e, a_t^e, s_t^p, a_t^p, r_t)$.
However, we only have $(y^0=x, y^*)$ in our dataset.
Traditional exploration methods like $\epsilon$-greedy sampling sample edits for all tokens independently, ignoring the interdependency between them.
Instead, we devise an \acf{IRT} algorithm to sample an edit for each token by considering its future expectation, i.e., sampling $a^e_{t, i}$ based on expectation of $a^e_{t, :i-1}$ from $i=|y^t|$ to $1$.
We maintain a matrix $M^t$ for this expectation based on both $y^t$ and $y^*$, which is computed by a \acf{DPS} process due to the exponential number of edit combinations of $a^e_{t, :i}$.
The details of \ac{IRT} are provided in Alg.~\ref{training}; it contains a \ac{DPS} process that consists of two parts: computing the matrix $M^t$~(line 4--8) and sampling actions $(a^e_t, a^p_t)$~(line 10) based on $M^t$.

\subsubsection{Computing the matrix $M^t$}
\begin{figure}[htb]
    \centering
    \includegraphics[width=0.49\textwidth]{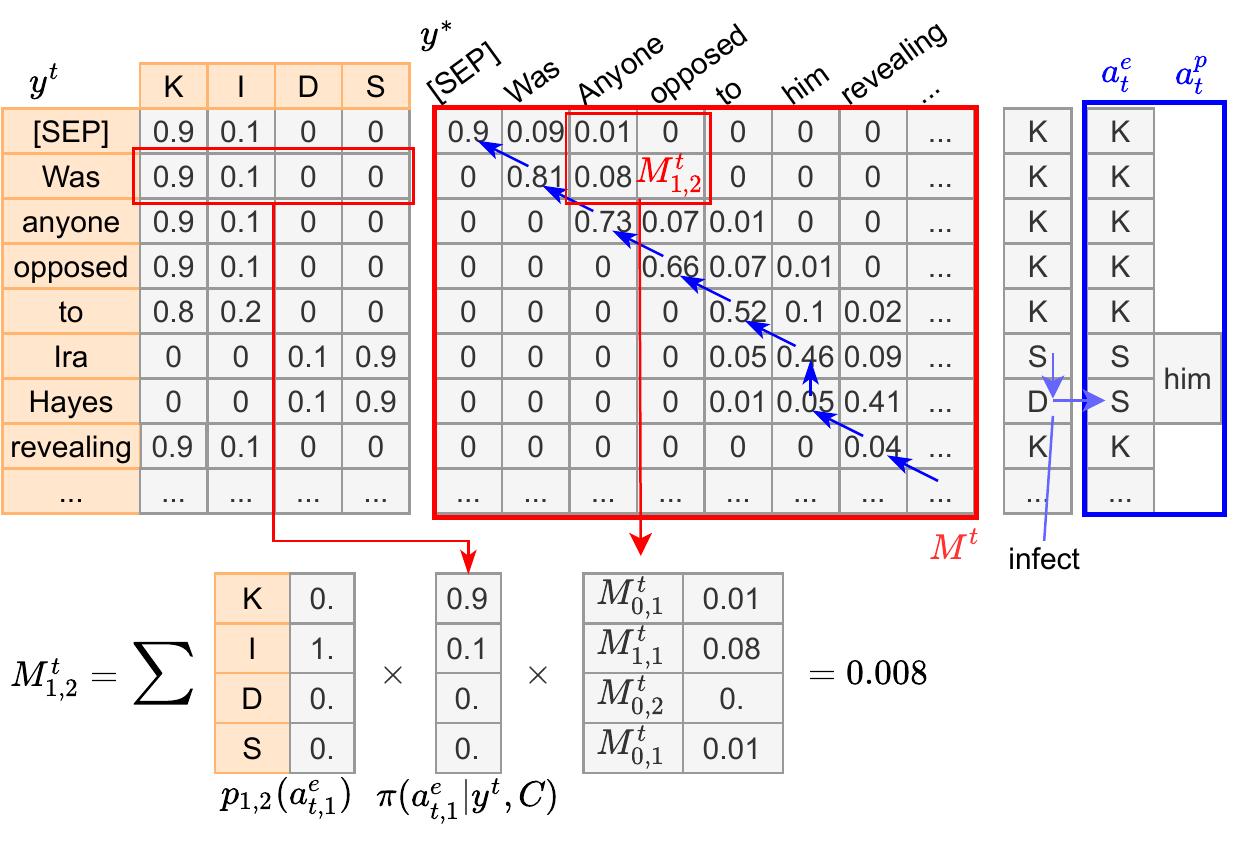}
    \caption{The \acs{DPS} process consists of computing matrix $M$~(red box) and sampling $(a^e_t, a^t_p)$~(blue arrows and box).}
    \label{fig:DPS}
\end{figure}

Given $(y^t, y^*)$ with length $m$ and $n$, we maintain a matrix $M^t \in \mathbb{R}^{(m + 1) \times (n + 1)}$ (including `[SEP]', see the upper right part in Fig.~\ref{fig:DPS}) where each element $M^t_{i, j}$ tracks the expectation of $a^e_{t, :i}$ to convert $y^t_{:i}$ to $y^*_{:j}$:
\begin{equation}
    \label{M_Expectation}
    \begin{split}
         M^t_{i, j} &= E_{p_{i, j}(a^e_{t, i})}[E_{p(a^e_{t, :i-1})} \pi_{y^t_{:i}->y^*_{:j}}(a^e_{t, :i})] \\
         &
         \begin{split}
            = {} & E_{p_{i, j}(a^e_{t, i})}\left [\pi^e(a^e_{t, i}|y^t, C) \times
            \vphantom{\begin{cases}X\\X\\X\\X\end{cases}}
            \right. \\
            & \left.\begin{cases}
            M^t_{i-1, j-1}, \text{ if } a^e_{t, i} = \text{`K'} \\
            M^t_{i-1, j}, \text{ if } a^e_{t, i} = \text{`D'} \\
            M^t_{i, j-1}, \text{ if } a^e_{t, i} = \text{`I'} \\
            M^t_{i-1, j-1}, \text{ if } a^e_{t, i} = \text{`S'}
            \end{cases}\!\!\right]\!,
        \end{split}
    \end{split}
\end{equation}
where $a^e_{t, :i}$ is the combinational edits for tokens $y^t_{:i}$ and $\pi^e(a^e_{t, i}|y^t, C)$ is calculated by Eq.~\ref{edit_prediction} (see the upper left part in Fig.~\ref{fig:DPS}).
$M^t_{0, 0}$ is initialized to $1$.
We will first introduce $p_{i, j}(a^e_{t, i})$ and then introduce $\pi_{y^t_{:i}->y^*_{:j}}(a^e_{t, :i})$ in Eq.~\ref{M_Expectation}.

Traditional sampling methods sample each edit $a^e_{t, i}$ independently, based on model likelihood $\pi^e(a^e_{t, i}|y^t, C)$.
Instead, we sample each edit with probability $p_{i, j}(a^e_{t, i})$ based on edits expectation $M^t$, which is modeled as:
\begin{equation}
    \label{p_computing}
    \begin{split}
    \mbox{}\hspace*{-2mm}
        p_{i, j}(a^e_{t, i}) ={} & \frac{1}{Z^t_{i, j}} \pi(a^e_{t, i} | y^t, C) \times \\
        & \begin{cases}
        M^t_{i-1, j-1}, \text{ if } a^e_{t, i} = \text{`K'} \\
        M^t_{i-1, j}, \text{ if }  a^e_{t, i} = \text{`D'} \\
        M^t_{i, j-1}, \text{ if }  a^e_{t, i} = \text{`I'} \\
        M^t_{i-1, j-1}, \text{ if }  a^e_{t, i} = \text{`S'},
        \end{cases}
    \end{split}
\end{equation}
where $Z^t_{i, j}$ is the normalization term.
We give an example on computing $M^t_{1, 2}$ in the bottom part of Fig.~\ref{fig:DPS}.
For edit `I' in $M^t_{1, 2}$, its probability is 1, and its value is $\pi^e(a^e_{t, i} = \text{`I'} | y^t, C) \times M^t_{1, 1}=0.008$.
For the other edits, the probability is 0.
Therefore, $M^t_{1, 2} = 0.008$.

$\pi_{y^t_{:i}->y^*_{:j}}(a^e_{t, :i})$ is the probability of conducting edits $a^e_{t, :i}$ to convert $y^t_{:i}$ to $y^*_{:j}$:
\begin{equation}
    \label{pi_factorization}
    \begin{split}
    \mbox{}\hspace*{-3mm}
        &\pi_{y^t_{:i}->y^*_{:j}}(a^e_{t, :i}) = \pi^e(a^e_{t, i}|y^t, C) \times \\
        &\begin{cases}
            \pi_{y^t_{:i - 1}->y^*_{:j - 1}} (a^e_{t, :i-1}), \text{ if } a^e_{t, i - 1} = \text{`K'} 
            \hspace*{-5mm}\mbox{}\\
            \pi_{y^t_{:i - 1}->y^*_{:j}} (a^e_{t, :i -1}), \text{ if } a^e_{t, i - 1} = \text{`D'} \\
            \pi_{y^t_{:i}->y^*_{:j - 1}} (a^e_{t, :i}), \text{ if } a^e_{t, i} = \text{`I'} \\
            \pi_{y^t_{:i - 1}->y^*_{:j - 1}} (a^e_{t, :i-1}), \text{ if } a^e_{t, i - 1} = \text{`S'},
            \hspace*{-5mm}\mbox{}
        \end{cases}
    \end{split}
\end{equation}
To convert $y^t_{:i}$ to $y^*_{:j}$, we need to make sure that $y^t_i$ can convert to $y^*_j$ and that $y^t_{:i-1}$ can convert to $y^*_{:j-1}$, which can be calculated recursively.
Note that we only allow `S' and `D' for $y^t_i$ when $y^t_i \neq y^*_j$ and `K' and `I' for $y^t_i$ when $y^t_i = y^*_j$.
And $M^t_{i-1,j-1} = E_{p(a^e_{t,:i-1})} \pi_{y^t_{:i-1}->y^*_{:j-1}}(a^e_{t, :i-1})$.

\begin{algorithm}[htb]
\SetAlgoLined
\KwInput{The origin data $\mathcal{D}=\{(x, y^*)\}$, the number of samples $L$;}
\KwOutput{The model parameters $\theta$;}
\While{not coverage}{
    Sample ($y^t$, $y^*$) from $\mathcal{D}$ \;
    $M^t_{0,0}=1$\;
    \For{i in \rm 0,\ldots, $m$}{
        \For{j in \rm 0,\ldots, $n$}{
            Compute $M^t_{i, j}$ according to Eq.~\ref{M_Expectation}\;
        }
    }
    Sample $a^{e}_t, a^{p}_t$ according to Eq.~\ref{ij_update} \; 
    Apply $a^{e}_t, a^{p}_t$ to obtain $y^{t + 1}$ \;
    Obtain $r_t$ according to Eq.~\ref{reward}  \;
    Update $\theta$ according to Eq.~\ref{PG} \;
    Add ($y^{t + 1}$, $y^*$) to $D$.
}
 \caption{Training Process of \ac{RISE}}
 \label{training}
\end{algorithm}

\subsubsection{Sampling $(a^e_t, a^p_t)$}
We sample $(a^e_t, a^p_t)$ based on matrix $M^t$ by backtracking from $i=m, j=n$.
For example, as shown in the upper right in Fig.~\ref{fig:DPS}, we backtrack along the blue arrows.
In this truncated sample, we start from $M^t_{7, 6}$, sample an edit `K' to keep `revealing' based on $p_{7, 6}(a^e_{t, 7})$ in Eq.~\ref{p_computing}, and move to $M^t_{6, 5}$.
Then, we sample `S' to substitute `Ira Hayes' to `him' and move to $M^t_{4, 4}$. 
Finally, we sample `K' in $[M^t_{4, 4}, M^t_{3, 3}, M^t_{2, 2} M^t_{1, 1}, M^t_{0, 0}]$ to keep [`to', `opposed', `anyone', `Was', `[SEP]'].
Therefore, we can obtain $a^e_t=$ [\ac{K}, \ac{K}, \ac{K}, \ac{K}, \ac{K}, \ac{S}, \ac{S}, \ac{K}], $a^p_t$ = [`him'].
Note that we obtain $a^{p}_t$ by merging all corresponding tokens $y^*_j$ as the phrase for each `I' edit and successive `S' edits and we only substitute once.
The backtracking rule can be formulated as:
\begin{equation}
    \label{ij_update}
    M^t_{i, j} \rightarrow
    \begin{cases} M^t_{i - 1, j - 1}, \text{if $a^e_{t, i} \in [\text{`K'}, \text{`S'}]$} \\ 
                M^t_{i - 1, j}, \text{if $a^e_{t, i} = \text{`D'}$} \\
                M^t_{i, j - 1}, \text{if $a^e_{t, i} = \text{`I'}$}.
    \end{cases}
\end{equation}

\subsection{Inference}
During inference, \ac{RISE} iteratively edits $x$ until it predicts `K' edits for all tokens or it achieves the maximum iteration limit.
For example, for editing iteration $t$ in Figure~\ref{fig:model}, it predicts `S' for `Ira' and `Hayes' to substitute it to `him' and `K' for other tokens, which results in `Was anyone opposed to him revealing \ldots' as output.
The output in iteration $t$ is the input of iteration $t+1$.
The actual editing iteration times vary with different samples.


\section{Experiments}

\begin{table*}[ht]
\centering
\caption{Overall performance (\%) on \acs{CANARD} and \acs{CAsT}. 
\textbf{Bold face} indicates the best results in terms of the corresponding metrics. 
Significant improvements over the best baseline results are marked with $^\ast$ (t-test, $p < 0.01$).
Note that we denote BLEU-$n$ as B-$n$ and ROUGE-L as R-L.}
\label{tab: result} 

\begin{tabular}{l c c c c c c c c c c c c}
\toprule
 & \multicolumn{6}{c}{\acs{CANARD}~(\%)} & \multicolumn{6}{c}{\acs{CAsT}~(\%) (unseen)} \\
 \cmidrule(r){2-7}\cmidrule{8-13}
\bf Method & \textbf{B-1} & \textbf{B-2} & \textbf{B-3} & \textbf{B-4}  & \textbf{R-L} & \textbf{CIDEr} & \textbf{B-1} & \textbf{B-2} & \textbf{B-3} & \textbf{B-4} & \textbf{R-L} & \textbf{CIDEr} \\ 
\midrule
Origin          & 54.7 & 47.0 & 40.6 & 35.3 & 70.9 & 3.460 & 75.9 & 69.2 & 62.9 & 57.6 & 85.0 & 5.946 \\
Rule            & 55.0 & 47.0 & 40.2 & 34.8 & 70.5 & 3.420 & 78.0 & 71.4 & 65.3 & 60.0 & 86.1 & 6.220 \\ 
\midrule
Trans++         & 84.3 & 77.5 & 72.1 & 67.5 & 84.6 & 6.348 & 76.0 & 64.3 & 54.8 & 47.2 & 76.5 & 4.258 \\ 
QGDiv           & 85.2 & 78.6 & 73.3 & 68.9 & 85.2 & 6.469 & 75.9 & 65.3 & 56.7 & 59.6 & 78.0 & 4.694 \\          
QuerySim        & 83.1 & 78.5 & 74.5 & 71.0 & 82.7 & 6.585 & 80.6 & 75.3 & 70.2 & 65.5 & 83.3 & 6.345 \\
\midrule
RISE            & \textbf{86.3}\rlap{$^\ast$} & \textbf{80.5}\rlap{$^\ast$} & \textbf{75.6} & \textbf{71.6}\rlap{$^\ast$} & \textbf{86.2}\rlap{$^\ast$} & \textbf{6.759}
                & \textbf{85.1}\rlap{$^\ast$} & \textbf{78.4} & \textbf{72.2} & \textbf{66.8} & \textbf{87.8}\rlap{$^\ast$} & \textbf{6.543}  \\  
\bottomrule
\end{tabular}

\end{table*}

\subsection{Datasets} \label{Datasets}
As with previous studies~\cite{elgohary2019can,conf/sigir/YuLYXBG020,vakulenko2020question,journals/corr/abs-2004-01909}, we conduct experiments on the \acs{CANARD}\footnote{\url{http://canard.qanta.org}}~\cite{elgohary2019can} dataset, which is a large open-domain dataset for conversational question answering~(with over 30k training samples).
Each sample in the \acs{CANARD} dataset includes a conversational context (historical questions and answers), an self-contained question, and its corresponding conversational question under the context.
The questions always have clear answers, e.g., `Did he win the lawsuit?'
We follow the \acs{CANARD} splits for training and evaluation.

In addition, we evaluate the model performance on the \acs{CAsT}\footnote{\url{http://www.treccast.ai}} dataset~\cite{dalton2019cast}, which is built for conversational search.
Different from \acs{CANARD}, its context only contains questions without corresponding answers.
Besides, most questions in the \acs{CAsT} dataset are exploring questions to explore relevant information, e.g., `What about for great whites?'
Since the \acs{CAsT} dataset only contains 479 samples from different domains compared to \acs{CANARD}, we use it for testing.


\subsection{Evaluation metrics}
Following \citet{conf/acl/SuSZSHNZ19,conf/emnlp/XuTSWZSY20}, we use BLEU-1, BLEU-2, BLEU-3, BLEU-4~\cite{conf/acl/PapineniRWZ02}, ROUGE-L~\cite{lin2004rouge}, and CIDEr~\cite{vedantam2015cider} for automatic evaluation.
BLEU-$n$ and ROUGE-L measure the word overlap between the generated and golden questions.
CIDEr measures the extent to which important information is missing.
\citet{elgohary2019can,journals/corr/abs-2004-01909,conf/emnlp/XuTSWZSY20} have shown that automatic evaluation has a high correlation with human judgement on this task, so we do not conduct human evaluation in this paper.

\subsection{Baselines}
We compare with several recent state-of-the-art methods for this task or closely related tasks:

\begin{itemize}[leftmargin=*,nosep]
\item \textbf{Origin} uses the original self-contained question as output.

\item \textbf{Rule}~\cite{conf/sigir/YuLYXBG020} employs two simple rules to mimic two conversational characteristics: anaphora and ellipsis.

\item \textbf{QGDiv}~\cite{conf/acl/SultanCAC20} uses RoBERTa~\cite{journals/corr/abs-1907-11692} with  beam search~\cite{conf/emnlp/WisemanR16} for generation.

\item \textbf{Trans++}~\cite{vakulenko2020question} predicts several word distributions, and combines them to obtain the final word distribution when generating each token.

\item \textbf{QuerySim}~\cite{conf/sigir/YuLYXBG020} adopts a GPT-2~\cite{radford2019language} model to generate conversational question. 
\end{itemize}
We also found some methods from related tasks.
But they do not work on this task for various reasons.
For example, due to the lack of labels needed for training, we cannot compare with the methods proposed by \citet{www/RossetXSCCTB20} and \citet{conf/emnlp/XuTSWZSY20}.
\citet{conf/acl/SuSZSHNZ19} propose a model that can only copy tokens from input;
it works well on the reverse task (i.e., \ac{CQR}), but not on \ac{CQS}. 

\subsection{Implementation details}
We use BERT2BERT for the modeling of the editing and phrasing parts~\cite{journals/tacl/RotheNS20}, as other pretrained models like GPT-2~\cite{radford2019language} cannot work for both.
The hidden size is 768 and phrase vocabulary is 3461 following~\cite{conf/emnlp/MalmiKRMS19}.
We use the BERT vocabulary~(30,522 tokens) for all BERT-based or BERT2BERT-based models. 
We use the Adam optimizer (learning rate 5e-5)~\cite{journals/corr/KingmaB14} to train all models.
In particular, we train all models for 20,000 warm-up steps, 5 epochs with pretrained model parameters frozen, and 20 epochs for all parameters.
For \ac{RISE}, the maximum editing iteration times is set to 3.
We use gradient clipping with a maximum gradient norm of 1.0. 
We select the best models based on the performance on the validation set. 
During inference, we use greedy decoding for all models.

\subsection{Results}
We list the results of all methods on both \acs{CANARD} and \acs{CAsT} in Table~\ref{tab: result}.
From the results, we have two main observations.

First, \ac{RISE} significantly outperforms all baselines on both datasets. 
Specifically, \ac{RISE} outperforms the strongest baseline QuerySim by \textasciitilde 4\% in terms of ROUGE-L.
The reason is that \ac{RISE} enhanced by \ac{DPS} has a better ability to emphasize conversational tokens, rather than treating all tokens equally.

Second, \ac{RISE} is more robust, which generalizes better to unseen data of \acs{CAsT}.
The results of the neural methods on \acs{CANARD} are much better than those on \acs{CAsT}.
But, \ac{RISE} is more stable than the other neural models.
For example, \ac{RISE} outperforms QuerySim by 0.6\% in BLEU-4 on \acs{CANARD}, while 1.3\% on \acs{CAsT}.
The reason is that \ac{RISE} learns to cope with conversational tokens only, while other models need to generate each token from scratch.





\section{Analysis}

\begin{table*}[t]
\centering
\caption{Ablation study (\%) on \acs{CANARD} and \acs{CAsT}. 
}
\label{tab: ablation_study}

\begin{tabular}{l c c c c c c  c c c c c c}
\toprule
 & \multicolumn{6}{c}{\acs{CANARD}~(\%)} & \multicolumn{6}{c}{\acs{CAsT}~(\%) (unseen)} \\
 \cmidrule(r){2-7}
  \cmidrule(r){8-13}
\bf Method & \textbf{B-1} & \textbf{B-2} & \textbf{B-3} & \textbf{B-4}  & \textbf{R-L} & \textbf{CIDEr} & \textbf{B-1} & \textbf{B-2} & \textbf{B-3} & \textbf{B-4} & \textbf{R-L} & \textbf{CIDEr} \\ 
\midrule
Origin          & 54.7 & 47.0 & 40.6 & 35.3 & 70.9 & 3.460 & 75.9 & 69.2 & 62.9 & 57.6 & 85.0 & 5.946 \\
\midrule
-DPS       & 67.5 & 56.4 & 47.3 & 39.9 & 73.9 & 3.743 & 80.9 & 70.0 & 60.6 & 53.3 & 81.2 & 4.713 \\
-MLD            & 85.2 & 78.6 & 73.3 & 68.9 & 85.2 & 6.469 & 75.9 & 65.3 & 56.7 & 59.6 & 78.0 & 4.694 \\
\midrule
RISE            & \textbf{86.3} & \textbf{80.5}\rlap{$^\ast$} & \textbf{75.6}\rlap{$^\ast$} & \textbf{71.6}\rlap{$^\ast$} & \textbf{86.2}\rlap{$^\ast$} & \textbf{6.759}\rlap{$^\ast$}
                & \textbf{85.1}\rlap{$^\ast$} & \textbf{78.4}\rlap{$^\ast$} & \textbf{72.2}\rlap{$^\ast$} & \textbf{66.8}\rlap{$^\ast$} & \textbf{87.8}\rlap{$^\ast$} & \textbf{6.543}\rlap{$^\ast$}  \\
\bottomrule
\end{tabular}

\end{table*}

\subsection{Ablation study}
To analyze where the improvements of \ac{RISE} come from, we conduct an ablation study on the \acs{CANARD} and \acs{CAsT} datasets~(see Table~\ref{tab: ablation_study}).
We consider two settings: 
\begin{itemize}[leftmargin=*,nosep] 


\item \textbf{-\acs{DPS}.} Here, we replace \ac{DPS} by $\epsilon$-greedy sampling~($\epsilon=0.2$) ~\citep{books/lib/SuttonB98}.

\item \textbf{-\ac{MLD}.} Here, we replace \ac{MLD} by \ac{MLE} in \ac{RISE}.

\end{itemize}

\noindent%
The results show that both parts (\ac{DPS} and \ac{MLD}) are helpful to \ac{RISE} as removing either of them leads to a decrease in performance.
Without \ac{MLD}, the performance drops a lot in terms of all metrics, e.g., 3\% and 7\% in BLEU-4 on \acs{CANARD} and \acs{CAsT}, respectively.
This indicates that optimizing \ac{MLD} is more effective than optimizing \ac{MLE}.  
Besides, \ac{MLD} generalizes better on unseen \acs{CAsT} as it drops slightly in all metrics, while with \ac{MLE}, we see a drop of 10\% in BLEU-1.

Without \ac{DPS}, the results drop dramatically, which indicates that \ac{DPS} can do better exploration than $\epsilon$-greedy and is of vital importance for \ac{RISE}.
For example, -\acs{DPS} tends to sample more non `K' edits~(\acs{RISE} vs -\acs{DPS}: 10\% vs 22\% on \acs{CANARD}), which is redundant and fragile.
The performance of -\ac{DPS} is even worse than Origin in \acs{CAsT} in BLEU-4.
This may be because \acs{CAsT} is unseen.



\subsection{Editing iterations}
\begin{figure}[tb]
    \centering
    \includegraphics[width=0.48\textwidth]{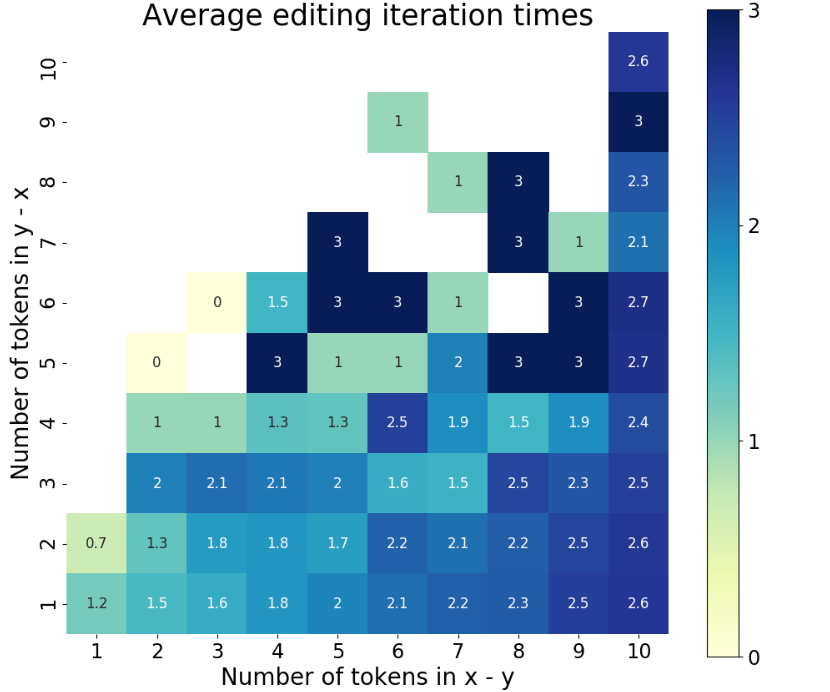}
    \caption{Average number of editing iteration of \ac{RISE} conditioned on number of tokens in $x$ - $y$ and $y$ - $x$.}
    \label{fig:edit_times}
\end{figure}

To analyze the relation between the number of editing iterations of \ac{RISE} and the editing difficulty, we plot a heatmap in Fig.~\ref{fig:edit_times}, where the deeper color represents a larger number of editing iterations.
The x-axis denotes the number of tokens shown in input $x$ but not shown in output $y$ and the y-axis denotes the number of tokens shown in $y$ but not in $x$.

As the number of different tokens between $x$ and $y$ increases, the number of editing iterations increases too.
For example, when the y-axis is 1, as the x-axis ranges from 1 to 10, the number of editing iterations increases from 1.2 to 2.6 because more `D' edits are needed.
We also found that when the x-axis is between 3 and 7 and the y-axis is between 1 and 4, only 1--2 editing iterations are needed.
Usually, this is because \ac{RISE} only needs 1 or 2 successive `S' edits for simulating anaphora.

\subsection{Influence of the number of editing iterations}
The overall performance of \ac{RISE} improves as the number of editing iterations increases.
\ac{RISE} achieves 70.5\% in BLEU-4 in the first iteration (even worse than QuerySim in Table~\ref{tab: result}) but 71.5\% and 71.6\% in the second and third iterations.
This shows that some samples are indeed more difficult to be directly edited into conversational ones, and thus need more editing iterations.

Even though it will not hurt the performance a lot, more editing iterations are not always helpful.
About 5\% of the samples achieve worse BLEU-4 scores as the number of editing iterations increases.
For example, \ac{RISE} edits `where did humphrey lyttelton go to school at?' into `where did he go to school at?' in the first iteration, which is perfect.
But \ac{RISE} continues to edit it into `where did he go to school?' in the second iteration, which is undesirable.
This is because \ac{RISE} fails to decide whether to stop or continue editing. 


\subsection{Case Study}
\begin{table}[t]
\centering
\caption{Examples generated by \ac{RISE} on \acs{CANARD}. Here, `Question' means the self-contained question, and `Target' means the desired conversational question. `Rewrite\#n' denotes the output of RISE in n-th iteration.}
\label{tab: case_study} 
\resizebox{\columnwidth}{!}{%
\begin{tabular}{l p{0.35\textwidth}}
\toprule
\textbf{Example~1} & 1.~At Tabuk the standard of the army was entrusted to Abu Bakr. \\ 
Context & 2.~Where was Tabuk located? \\ 
 & 3.~Tabuk on the Syrian border. \\
\midrule
Question   &  What did Abu Bakr do during the expedition of Tabuk? \\
Rewrite\#1  & What did he bakr do during expedition? \\
Rewrite\#2  & What did he do during expedition? \\
Target   &  What did abu bakr do during the expedition? \\
\midrule
\textbf{Example~2} & 1.~When did Clift start his film career? \\
Context & 2.~His first movie role was opposite John Wayne in Red River, which was shot in 1946 and released in 1948. \\
\midrule
Question   &  Did Montgomery Clift win any awards for any of his films? \\
Rewrite\#1  & Did he win any awards for and? \\
Rewrite\#2  & Did he win any awards? \\
Target   &  Did he win any awards for any of his films? \\
\bottomrule
\end{tabular}}
\end{table}

In Table~\ref{tab: case_study} we present two examples of the output of RISE.
We present the context, the original self-contained question, the target conversational question, and the output of RISE in the $n$-th iteration, denoted as `Context', `Question', `Target' and `Rewrite\#n', respectively.
We have two main observations.
First, it is helpful to edit iteratively.
As shown in Example 1, RISE first replaces `Abu' as `he' in the first iteration and then deletes `bakr' in the second iteration, which simulates anaphora by editing twice.
In Example 2, RISE simulates ellipsis by deleting multiple words and achieves poor grammar after the first iteration but corrects this by deleting some of the leftover words.
RISE may have learned to check the grammar and remove redundant words.

Second, RISE can simulate more conversational characteristics than human, and sometimes it can achieve a better result, sometimes not.
As we can see, RISE results a better conversational question by additionally simulating anaphora for `Abu Bakr' in Example 1.
However, RISE leaves out necessary information in Example 2.
Here, RISE tries to simulate conversational characteristics as much as possible, where the result may be uncontrollable.
In future work, we will add a discriminator to check the necessary information.


\section{Related work}
Studies on asking conversational question can be divided into two categories: \emph{conversational question generation} and \emph{conversational question ranking}.

Conversational question generation aims to directly generate conversational questions conditioned on the dialogue context~\cite{conf/acl/SultanCAC20,RenTois}.
\citet{conf/www/ZamaniDCBL20} and \citet{conf/emnlp/0003ZM20} define a question utility function to guide the generation of conversational questions.
\citet{conf/acl-mrqa/NakanishiKH19,conf/acl/JiaZSW20} incorporate knowledge with auxiliary tasks.
These methods may generate irrelevant questions due to their pure generation nature.

Conversational question ranking~\cite{sigir/AliannejadiZCC19} retrieves questions from a collection based on the given context, so the questions are mostly relevant to the context.
\citet{conf/acl/KunduLN20} propose a pair-wise matching network between context and question to do question ranking.
Some studies also use auxiliary tasks to improve ranking performance, such as \acl{NLI}~\cite{conf/cikm/KumarRC20} and relevance classification~\cite{www/RossetXSCCTB20}.
The retrieved questions are often unnatural without considering the conversational characteristics, e.g., anaphora and ellipsis.

\ac{CQS} rewrites the retrieved self-contained questions into conversational ones by incorporating the conversational characteristics.
Existing applicable methods for \ac{CQS} are all \ac{MLE} based~\cite{conf/emnlp/XuTSWZSY20,conf/sigir/YuLYXBG020,journals/corr/abs-2005-02230,vakulenko2020question}, which often get stuck in easily learned tokens as each token is treated equally by \ac{MLE}.
Instead, we propose a \ac{MLD} based \ac{RISE} framework to formulate \ac{CQS} as a \ac{HCMDP}, which is able to discriminate different tokens through explicit editing actions, so that it can learn to emphasize the conversational tokens and generate more natural and appropriate questions.


\section{Conclusion}
In this paper, we have proposed a \acf{MLD} based \acf{RISE} framework for \acf{CQS}.
To train \ac{RISE}, we have devised an \acf{IRT} algorithm with a novel \acf{DPS} process.
Extensive experiments show that \acs{RISE} is more effective and robust than several state-of-the-art \ac{CQS} methods. 
A limitation of \acs{RISE} is that it may fail to decide whether to stop or continue editing and leave out necessary information. 
In future work, we plan to address this issue by learning a reward function that considers the whole editing process through adversarial learning~\cite{journals/corr/GoodfellowPMXWOCB14}.

\section*{Code}
To facilitate the reproducibility of the results, we share the codes of all methods at \url{https://github.com/LZKSKY/CaSE_RISE}.

\section*{Acknowledgments}
We thank the reviewers for their valuable feedback.
This research was partially supported by
the National Key R\&D Program of China with grant No. 2020YFB1406704,
the Natural Science Foundation of China (61972234, 61902219, 62072279), 
the Key Scientific and Technological Innovation Program of Shandong Province (2019JZZY010129), 
the Tencent WeChat Rhino-Bird Focused Research Program (JR-WXG-2021411),
the Fundamental Research Funds of Shandong University,
and 
the Hybrid Intelligence Center, a 10-year program funded by the Dutch Ministry of Education, Culture and Science through the Netherlands Organisation for Scientific Research, \url{https://hybrid-intelligence-centre.nl}.

All content represents the opinion of the authors, which is not necessarily shared or endorsed by their respective employers and/or sponsors.

\bibliographystyle{acl_natbib}

\clearpage
\section*{Appendix}
For reproducibility for all reported experimental results, we report the following information.
The average running time for RISE, QuerySim, Trans++, QGDiv, -MLD, -DPS are 15 hours, 5 hours, 9.5 hours, 9 hours, 9 hours, 15 hours, respectively.
The number of parameters in RISE, Trans++, QGDiv, -MLD, -DPS are 221M and the number of parameters in QuerySim is 125M.
We list the validation performance on \acs{CANARD} in Table.~\ref{result_valid}, as only \acs{CANARD} is used for validation.
As we can see, it has high correlation to test performance on \acs{CANARD}.
We use this script~\footnote{https://github.com/Maluuba/nlg-eval} for evaluation.

\begin{table}[htbp]
\centering
\caption{Overall performance (\%) on validation set of \acs{CANARD}.
Note that we denote BLEU-n as B-n and ROUGE-L as R-L.}

\label{result_valid}

\setlength{\tabcolsep}{1.3mm}{
\begin{tabular}{l c c c c c c}
\toprule
 & \multicolumn{6}{c}{\acs{CANARD}~(\%)} \\
\midrule
\bf Method & \textbf{B-1} & \textbf{B-2} & \textbf{B-3} & \textbf{B-4}  & \textbf{R-L} & \textbf{CIDEr}  \\ 
\midrule
Trans++         & 86.5 & 80.3 & 75.4 & 71.3 & 86.2 & 6.704 \\ 
QGDiv           & 87.0 & 80.9 & 75.9 & 61.8 & 86.8 & 6.786 \\          
QuerySim        & 83.9 & 79.7 & 75.9 & 72.5 & 83.2 & 6.737 \\
\midrule
-DPS            & 67.2 & 55.9 & 46.8 & 39.4 & 74.3 & 3.745 \\ 
-MLD            & 87.0 & 80.9 & 75.9 & 61.8 & 86.8 & 6.786 \\ 
RISE            & 88.0 & 82.6 & 78.3 & 74.6 & 87.5 & 7.050 \\  
\bottomrule
\end{tabular}

}
\end{table}

For reproducibility for experiments with hyperparameter search, we report the following information.
The hyperparameter for \acs{RISE} is the max editing iteration times. We search it in range of 1 to 5 and find 3 can perform best on BLEU-4.
The results in range of 1 to 5 on BLEU-4 are 70.5\%, 71.5\%, 71.6\%, 71.6\% and 71.6\%, respectively.

\end{document}